\documentclass{article}

\usepackage[nonatbib,preprint]{neurips_2020}

\usepackage{graphicx}
\graphicspath{{./figs/}}
\usepackage{amsmath,amsfonts,bm,amssymb}

\def\eqref#1{equation~\ref{#1}}
\def\1{\bm{1}}

\DeclareMathAlphabet{\mathsfit}{\encodingdefault}{\sfdefault}{m}{sl}
\SetMathAlphabet{\mathsfit}{bold}{\encodingdefault}{\sfdefault}{bx}{n}

\def\gA{{\mathcal{A}}}

\def\gD{{\mathcal{D}}}

\def\gH{{\mathcal{H}}}

\def\gN{{\mathcal{N}}}
\def\gO{{\mathcal{O}}}

\def\gS{{\mathcal{S}}}
\def\gT{{\mathcal{T}}}

\newcommand{\E}{\mathbb{E}}

\newcommand{\R}{\mathbb{R}}

\newcommand{\KL}{D_{\mathrm{KL}}}

\usepackage[utf8]{inputenc} % allow utf-8 input
\usepackage[T1]{fontenc}    % use 8-bit T1 fonts
\usepackage{hyperref}       % hyperlinks
\usepackage{url}            % simple URL typesetting
\usepackage{booktabs}       % professional-quality tables
\usepackage{amsfonts}       % blackboard math symbols
\usepackage{nicefrac}       % compact symbols for 1/2, etc.
\usepackage{microtype}      % microtypography
\usepackage[dvipsnames]{xcolor}
\usepackage[utf8]{inputenc}
\usepackage[english]{babel}
\usepackage{subfig}

\usepackage{amsthm}
\usepackage{amsmath}
\usepackage{caption}
\usepackage[capitalise,nameinlink,noabbrev]{cleveref}
\usepackage{algorithm}
\usepackage{algorithmicx}
\usepackage{algpseudocode}
\usepackage{wrapfig}
\usepackage{multirow}

\algnewcommand{\LeftCommentX}[1]{\Statex \(\triangleright\) #1}
\algnewcommand{\LeftComment}[1]{\State \(\triangleright\) #1}
\makeatother

\theoremstyle{definition}

\title{Image Augmentation Is All You Need: \\ Regularizing Deep Reinforcement Learning  \\ from Pixels}

\author{
  Ilya Kostrikov\thanks{Equal contribution. Author ordering determined by coin flip. Both authors are corresponding.} \\
  New York University \\
  \texttt{kostrikov@cs.nyu.edu} \\
  \And
  Denis Yarats\footnotemark[1]\\
  New York University \& Facebook AI Research \\
  \texttt{denisyarats@cs.nyu.edu} \\
  \And
  Rob Fergus \\
  New York University \\
  \texttt{fergus@cs.nyu.edu} \\
}

\newcommand{\drq}{{\bf DrQ} }

\newcommand{\comment}[1]{}

\definecolor{target_q_aug}{rgb}{0.43921568627,0.67843137254,0.27843137254}
\definecolor{q_aug}{rgb}{0.26666666666,0.44705882352,0.76862745098}
\definecolor{f_func}{rgb}{1.0, 0.49019607843, 0.19215686274}

\begin{document}

\maketitle

\newif\ifincludeappendix

\begin{abstract}
  We propose a simple data augmentation technique that can be applied to standard model-free reinforcement learning algorithms, enabling robust learning directly from pixels without the need for auxiliary losses or pre-training. The approach leverages input perturbations commonly used in computer vision tasks to transform input examples, as well as regularizing the value function and policy. Existing model-free approaches, such as Soft Actor-Critic (SAC)~\cite{haarnoja2018sac}, are not able to train deep networks effectively from image pixels. However, the addition of our augmentation method dramatically improves SAC's performance, enabling it to reach state-of-the-art performance on the DeepMind control suite, surpassing model-based~\cite{hafner2019dream, lee2019slac, hafner2018planet} methods and recently proposed contrastive learning~\cite{srinivas2020curl}. Our approach, which we dub \textbf{DrQ}: \textbf{D}ata-\textbf{r}egularized \textbf{Q}, can be combined with any model-free reinforcement learning algorithm. We further demonstrate this by applying it to DQN~\cite{mnih2013dqn} and significantly improve its data-efficiency on the Atari 100k~\cite{kaiser2018simple} benchmark. An implementation can be found at \url{https://sites.google.com/view/data-regularized-q}.
\end{abstract}

\section{Introduction}
\begin{figure}[t!]
    \centering
    %\vspace{-20pt}
    \subfloat[Unmodified SAC.]{\includegraphics[width=1.\linewidth]{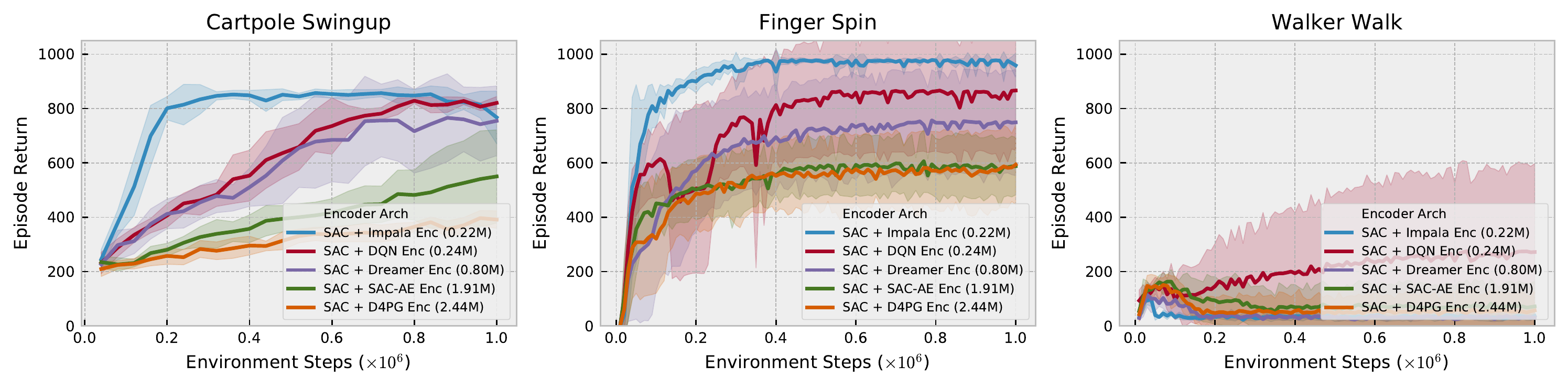}  \label{fig:encoders_a}}\\
    %\vspace{-10pt}
    \subfloat[SAC with image shift augmentation.]{\includegraphics[width=1.\linewidth]{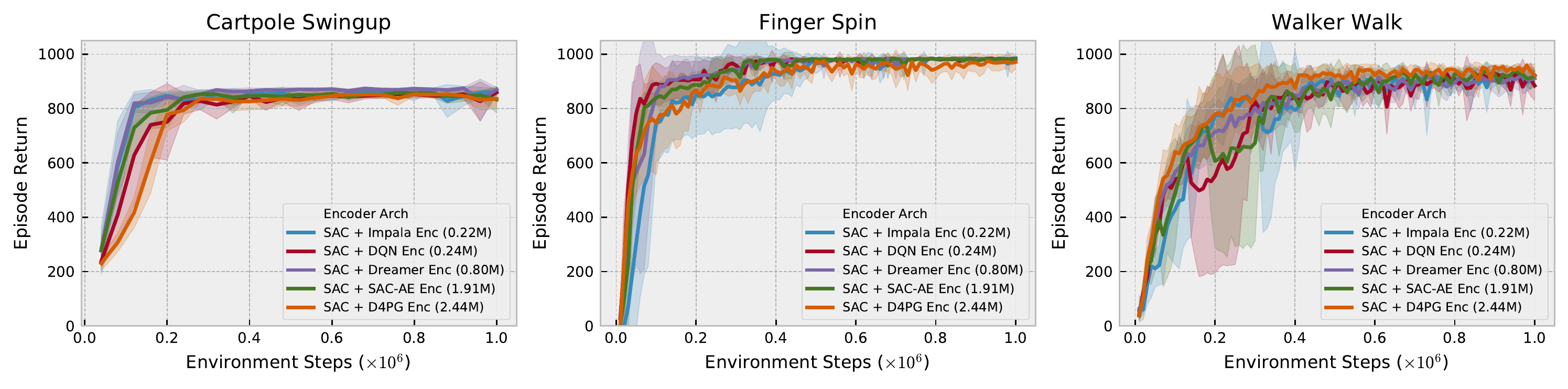} \label{fig:encoders_b}}
    \caption{The performance of SAC trained from pixels on the DeepMind control suite using image encoder networks of different capacity (network architectures taken from recent RL algorithms, with parameter count indicated). {\bf (a)}: unmodified SAC. Task performance can be seen to get {\em worse} as the capacity of the encoder increases, indicating over-fitting. For Walker Walk (right), all architectures provide mediocre performance, demonstrating the inability of SAC to train directly from pixels on harder problems. {\bf (b)}: SAC combined with image augmentation in the form of random shifts. The task performance is now similar for all architectures, regardless of their capacity. There is also a clear performance improvement relative to (a), particularly for the more challenging Walker Walk task. } 
    \label{fig:encoders}
    %\vspace{-10pt}
\end{figure}

Sample-efficient deep reinforcement learning (RL) algorithms capable of directly training from image pixels would open up many real-world applications in control and robotics. However, simultaneously training a convolutional encoder alongside a policy network is challenging when given limited environment interaction, strong correlation between samples and a typically sparse reward signal. Naive attempts to use a large capacity encoder result in severe over-fitting (see~\cref{fig:encoders_a}) and smaller encoders produce impoverished representations that limit task performance.

%\todor{lets fix paragraphs 2-3 here}

%\rob{Limited experience, small training}

Limited supervision is a common problem across AI and a number of approaches are adopted: (i) pre-training with self-supervised learning (SSL), followed by standard supervised learning; (ii) supervised learning with an additional auxiliary loss and (iii) supervised learning with data augmentation. 
SSL approaches are highly effective in the large data regime, e.g. in domains such as vision~\cite{chen2020simple,he2019moco} and NLP ~\cite{collobert2011natural,devlin2018bert} where large (unlabeled) datasets are readily available. However, in sample-efficient RL, training data is more limited due to restricted interaction between the agent and the environment, resulting in only $10^4$--$10^5$ transitions from a few hundred trajectories. While there are concurrent efforts exploring SSL in the RL context \cite{srinivas2020curl}, in this paper we take a different approach, focusing on data augmentation.%, limiting the effectiveness of SSL methods. 

A wide range of auxiliary loss functions have been proposed to augment supervised objectives, e.g. weight regularization, noise injection \cite{hinton2012improving}, or various forms of auto-encoder \cite{kingma2014semi}. In RL, reconstruction objectives \cite{jaderberg2016unreal,yarats2019improving} or alternate tasks are often used \cite{dwibedi2018visrepr}. However, these objectives are unrelated to the task at hand, thus have no guarantee of inducing an appropriate representation for the policy network. 

Data augmentation methods have proven highly effective in vision and speech domains, where output-invariant perturbations can easily be applied to the labeled input examples. Surprisingly, data augmentation has received relatively little attention in the RL community, and this is the focus of this paper. 
% From abstract: "The approach3leverages input perturbations commonly used in computer vision tasks to transforminput examples, as well as regularizing the value function and policy. "
The key idea is to use standard image transformations to peturb input observations, as well as regularizing the $Q$-function learned by the critic so that different transformations of the same input image have similar $Q$-function values. No further modifications to standard actor-critic algorithms are required, obviating the need for additional losses, e.g. based on auto-encoders \cite{yarats2019improving}, dynamics models \cite{hafner2018planet,hafner2019dream}, or contrastive loss terms \cite{srinivas2020curl}.

The paper makes the following contributions: (i) we demonstrate how straightforward image augmentation, applied to pixel observations, greatly reduces over-fitting in sample-efficient RL settings, without requiring any change to the underlying RL algorithm. (ii) exploiting MDP structure, we introduce two simple mechanisms for regularizing the value function which are generally applicable in the context of model-free off-policy RL. (iii) Combined with vanilla SAC \cite{haarnoja2018sac} and using hyper-parameters fixed across all tasks, the overall approach obtains state-of-the-art performance on the DeepMind control suite \cite{tassa2018dmcontrol}. (iv) Combined with a DQN-like agent, the approach also obtains state-of-the-art performance on the Atari 100k benchmark. (v) It is thus the first effective approach able to train directly from pixels without the need for unsupervised auxiliary losses or a world model. (vi) We also provide a PyTorch implementation of the approach combined with SAC and DQN.

\section{Background}

\textbf{Reinforcement Learning from Images} We formulate image-based control as an infinite-horizon partially observable Markov decision process (POMDP) \cite{bellman1957mdp,kaelbling1998planning}. An POMDP can be described as the tuple $(\gO, \gA, p, r, \gamma)$, where $\gO$ is the high-dimensional observation space (image pixels), $\gA$ is the action space, the transition dynamics $p = Pr(o'_{t}|o_{\leq t},a_{t})$ capture the probability distribution over the next observation $o'_t$ given the history of previous observations $o_{\leq t}$ and current action $a_{t}$, $r: \gO \times \gA \rightarrow \R$ is the reward function that maps the current observation and action to a reward $r_t  = r(o_{\leq t}, a_t)$, and $\gamma \in [0, 1)$ is a discount factor. Per common practice~\cite{mnih2013dqn}, throughout the paper the POMDP is converted into an MDP~\cite{bellman1957mdp} by stacking several consecutive image observations into a state $s_t = \{o_t, o_{t-1}, o_{t-2}, \ldots\}$. For simplicity we redefine  the transition dynamics $p = Pr(s'_{t}|s_t,a_{t})$ and the reward function $r_t = r(s_t, a_t)$. We then aim to find a policy $\pi(a_t|s_t)$ that maximizes the cumulative discounted return $\E_{\pi}[\sum_{t=1}^{\infty} \gamma^t r_t | a_t \sim \pi(\cdot|s_t), s'_t \sim p(\cdot|s_t, a_t), s_1 \sim p(\cdot)]$.

\textbf{Soft Actor-Critic} 
The Soft Actor-Critic (SAC)~\cite{haarnoja2018sac} learns a state-action value function $Q_\theta$, a stochastic policy $\pi_\theta$  and a temperature $\alpha$ to find an optimal policy for an MDP $(\gS, \gA, p, r, \gamma)$ by optimizing a $\gamma$-discounted maximum-entropy objective~\cite{ziebart2008maxent}.
$\theta$ is used generically to denote the parameters updated through training in each part of the model.

\textbf{Deep Q-learning} DQN~\cite{mnih2013dqn} also learns a convolutional neural net to approximate Q-function over states and actions. The main difference is that DQN operates on discrete actions spaces, thus the policy can be directly inferred from Q-values.
In practice, the standard version of DQN is frequently combined with a set of refinements that improve performance and training stability, commonly known as Rainbow~\cite{hasselt2015doubledqn}.
For simplicity, the rest of the paper describes a generic actor-critic algorithm rather than  DQN or SAC in particular. Further background on DQN and SAC can be found in~\cref{section:extended_background}.

\section{Sample Efficient Reinforcement Learning from Pixels}

This work focuses on the data-efficient regime, seeking to optimize performance given limited environment interaction. %However, with images as input a convnet encoder must be trained in conjunction with the policy network.
In \cref{fig:encoders_a} we show a motivating experiment that demonstrates over-fitting to be a significant issue in this scenario.  Using three tasks from the DeepMind control suite \cite{tassa2018dmcontrol}, SAC~\cite{haarnoja2018sac} is trained with the same policy network architecture but using different image encoder architectures, taken from the following RL approaches: NatureDQN~\cite{mnih2013dqn}, Dreamer~\cite{hafner2019dream}, Impala~\cite{espeholt2018impala}, SAC-AE~\cite{yarats2019improving} (also used in CURL~\cite{srinivas2020curl}), and D4PG~\cite{barth-maron2018d4pg}. The encoders vary significantly in their capacity, with parameter counts ranging from 220k to 2.4M. The curves show that {\em performance  decreases as parameter count increases}, a clear indication of over-fitting. 

\subsection{Image Augmentation}
\label{section:image_shift}
A range of successful image augmentation techniques to counter over-fitting have been developed in computer vision ~\cite{ciregan2012multi, cirecsan2011high, simard2003best, krizhevsky2012imagenet, chen2020simple}. These apply transformations to the input image for which the task labels are invariant, e.g. for object recognition tasks, image flips and rotations do not alter the semantic label. 
However, tasks in RL differ significantly from those in vision and in many cases the reward would not be preserved by these transformations. We examine several common image transformations from~\cite{chen2020simple} in~\cref{section:aug_ablation} and conclude that random shifts strike a good balance between simplicity and performance, we therefore limit our choice of augmentation to this transformation.

\cref{fig:encoders_b} shows the results of this augmentation applied during SAC training. We apply data augmentation only to the images sampled from the replay buffer and not for samples collection procedure.
The images from the DeepMind control suite are $84\times 84$. We pad each side by $4$ pixels (by repeating boundary pixels) and then select a random $84\times 84$ crop, yielding the original image shifted by $\pm 4$ pixels. This procedure is repeated every time an image is sampled from the replay buffer.
The plots show overfitting is greatly reduced, closing the performance gap between the encoder architectures. These random shifts alone enable SAC to achieve competitive absolute performance, without the need for auxiliary losses.

\subsection{Optimality Invariant Image Transformations}

\label{section:target_aug}
While the image augmentation described above is effective, it does not fully exploit the MDP structure inherent in RL tasks. We now introduce a general framework for regularizing the value function through transformations of the input state. 
For a given task, we define an optimality invariant state transformation $f: \mathcal{S} \times \gT \rightarrow \mathcal{S}$ as a mapping that preserves the $Q$-values
\begin{align*}
   Q(s, a) &= Q(f(s, \nu), a) \mbox{ for all } s \in \mathcal{S}, a \in \mathcal{A} \mbox{ and } \nu \in \gT.
\end{align*}
where $\nu$ are the parameters of $f(\cdot)$, drawn from the set of all possible parameters $\gT$.
One example of such transformations are the random image translations successfully applied in the previous section. 

For every state, the transformations allow the generation of several surrogate states with the same $Q$-values, thus providing a mechanism to reduce the variance of $Q$-function estimation. In particular, for an arbitrary distribution of states $\mu(\cdot)$ and policy $\pi$, instead of using a single sample $s^{*} \sim \mu(\cdot)$, $a^{*} \sim \pi(\cdot | s^{*})$ estimation of the following expectation
\begin{align*}
    \E_{\substack{s \sim \mu(\cdot)\\a \sim \pi(\cdot | s)}} [Q(s, a)] &\approx Q(s^{*}, a^{*})
\end{align*}
we can instead generate $K$ samples via random transformations and obtain an estimate with lower variance
\begin{align*}
\E_{\substack{s \sim \mu(\cdot)\\a \sim \pi(\cdot | s)}}[Q(s,a)] &\approx \frac{1}{K} \sum_{k=1}^{K} Q(f(s^{*}, \nu_k), a_k) \mbox{ where } \nu_k \in \gT \mbox{ and } a_k \sim \pi(\cdot|f(s^{*}, \nu_k)).
\end{align*}
\comment{        
\begin{itemize}
    \item We show that this idea can be implemented in several ways and provides some improvement.
\end{itemize}
}

This suggests two distinct ways to regularize $Q$-function. First, we use the data augmentation to compute the target values for every transition tuple $(s_i, a_i, r_i, s'_i)$ as
\begin{align}
    y_i& = r_i + \gamma \cfrac{1}{K}\sum_{k=1}^{K} Q_{\theta}(f(s'_i, \nu'_{i,k}), a'_{i, k}) \mbox { where } a'_{i, k} \sim \pi(\cdot | f(s'_i, \nu'_{i, k}))
    \label{eqn:y_target}
\end{align}
where $\nu'_{i,k} \in \gT$ corresponds to a transformation parameter of $s'_i$. Then the Q-function is updated using these targets through an SGD update using  learning rate $\lambda_\theta$
\begin{align}
\theta &\leftarrow \theta - \lambda_{\theta} \nabla_\theta \frac{1}{N} \sum_{i=1}^{N} (Q_{\theta}(f(s_i, \nu_i), a_i) - y_i)^2.
\label{eqn:targetq_update}
\end{align}

In tandem, we note that the same target from~\cref{eqn:y_target} can be used for different augmentations of $s_i$, resulting in the second regularization approach
\begin{align}
\theta &\leftarrow \theta - \lambda_{\theta} \nabla_\theta \frac{1}{NM} \sum_{i=1,m=1}^{N,M} (Q_{\theta}(f(s_i, \nu_{i,m}), a_i) - y_i)^2.
\label{eqn:q_update}
\end{align}

When both regularization methods are used,  $\nu_{i,m}$ and $\nu'_{i,k}$ are drawn independently.

\subsection{Our approach: Data-regularized Q (DrQ)}
Our approach, \textbf{DrQ}, is the union of the three separate regularization mechanisms introduced above: 
%(i) transformations of the input image (\cref{section:image_shift}); (ii) averaging the $Q$ target over \textsc{K} image transformations (\cref{eqn:y_target}) and (iii) averaging the $Q$ function itself over \textsc{M} image transformations (\cref{eqn:q_update}).
\begin{enumerate}
    \item transformations of the input image (\cref{section:image_shift}).
    \item averaging the $Q$ target over \textsc{K} image transformations (\cref{eqn:y_target}).
    \item averaging the $Q$ function itself over \textsc{M} image transformations (\cref{eqn:q_update}).
\end{enumerate}
\cref{alg:final} details how they are incorporated into a generic pixel-based off-policy actor-critic algorithm. If \textsc{[K=1,M=1]} then \drq reverts to \emph{image transformations alone}, this makes applying \drq to any model-free RL algorithm straightforward as it does not require any modifications to the algorithm itself. Note that \drq \textsc{[K=1,M=1]} also exactly recovers the concurrent work of RAD~\cite{laskin2020rad}, up to a particular choice of hyper-parameters and data augmentation type. 

\begin{figure}[ht!]
    \centering
    %\vspace{-20pt}
    \includegraphics[width=1.\linewidth]{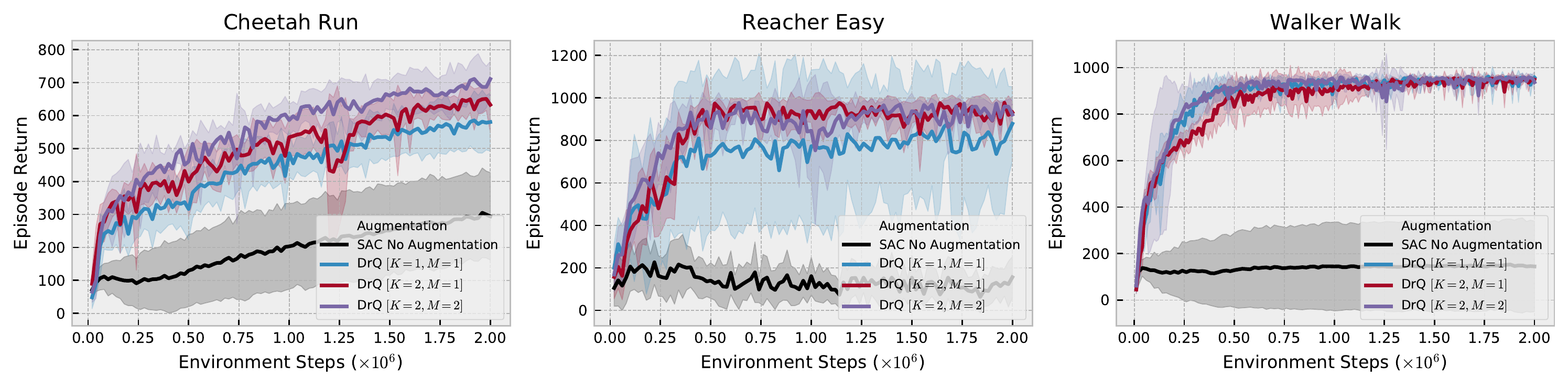}
    \caption{Different combinations of our three regularization techniques on tasks from \cite{tassa2018dmcontrol} using SAC.  Black: standard SAC. Blue: \textbf{DrQ} \textsc{[K=1,M=1]}, SAC augmented with random shifts. Red: \drq \textsc{[K=2,M=1]}, random shifts + Target Q augmentations. Purple: \textbf{DrQ} \textsc{[K=2,M=2]}, random shifts + Target Q + Q augmentations. All three regularization methods correspond to Algorithm 1 with different hyperparameters \textsc{K,M} and independently provide beneficial gains over unaugmented SAC. Note that \textbf{DrQ} \textsc{[K=1,M=1]} exactly recovers the concurrent work of RAD~\cite{laskin2020rad} up to a particular choice of hyper-parameters and data augmentation type. }
    \label{fig:versions}
\end{figure}

For the experiments in this paper, we pair \textbf{DrQ} with SAC~\cite{haarnoja2018sac} and DQN~\cite{mnih2013dqn}, popular model-free algorithms for control in continuous and discrete action spaces respectively. We select image shifts as the class of image transformations $f$, with $\nu \pm 4$, as explained in \cref{section:image_shift}. For target Q and Q augmentation we use \textsc{[K=2,M=2]} respectively.~\cref{fig:versions} shows \textbf{DrQ} and ablated versions, demonstrating clear gains over unaugmented SAC.  A more extensive ablation can be found in~\cref{section:km}.

\begin{algorithm*}[ht!]
\caption{
\textbf{DrQ}: \textbf{D}ata-\textbf{r}egularized \textbf{Q} applied to a generic off-policy actor critic algorithm.\\
Black: unmodified off-policy actor-critic.\\ {\color{f_func} Orange: image transformation}.\\ {\color{target_q_aug} Green: target $Q$ augmentation}.\\ {\color{q_aug} Blue: $Q$ augmentation}.
}

\begin{algorithmic}
  \State \textbf{Hyperparameters:}
    Total number of environment steps
    $T$,
    mini-batch size $N$,
    learning rate $\lambda_\theta$,
    target network update rate $\tau$,
    {\color{f_func} image transformation $f$},
    {\color{target_q_aug}
    number of target $Q$ augmentations $K$},
    {\color{q_aug}
    number of $Q$ augmentations $M$}.
    \For{each timestep $t=1..T$}
    \State  $a_t \sim \pi(\cdot | s_{t})$
    \State  $s'_t \sim p(\cdot | s_{t}, a_t)$
    \State $\mathcal{D} \leftarrow \mathcal{D} \cup (s_t, a_t, r(s_t, a_t), s'_t)$
    \State \textsc{UpdateCritic}($\mathcal{D}$)
    \State \textsc{UpdateActor}($\mathcal{D}$)
    \Comment Data augmentation is applied to  the samples for actor training as well.
    \EndFor
    \Procedure{UpdateCritic}{$\mathcal{D}$}
        \State $\{(s_i, a_i, r_i, s'_i)\}_{i=1}^{N} \sim \mathcal{D}$
        \Comment Sample a mini batch
        \State \textcolor{target_q_aug}{$\left\{ \nu'_{i,k}\middle| \nu'_{i,k} \sim \mathcal{U}(\gT), i=1..N, k=1..K \right\}$
        \Comment Sample parameters of target augmentations}
        \For{each $i=1..N$}
            \State $ a_i' \sim \pi(\cdot | s_i') $ \mbox { or } \textcolor{target_q_aug}{$ a'_{i,k} \sim \pi(\cdot | $} \textcolor{f_func}{$f$}\textcolor{target_q_aug}{$(s_i', \nu'_{i,k}))$\mbox{,  }$k=1..K$
            }
            \State $\hat{Q}_{i} = Q_{\theta'}(s'_i, a'_{i})$ \mbox{ or } \textcolor{target_q_aug}{$\hat{Q}_{i} = \frac{1}{K}\sum_{k=1}^KQ_{\theta'}($}\textcolor{f_func}{$f$}\textcolor{target_q_aug}{$(s'_i, \nu'_{i,k}), a'_{i,k})$ }
            \State $y_i \leftarrow r(s_i, a_i) + \gamma  \hat{Q}_i$
        \EndFor
        \State \textcolor{q_aug}{$\left\{ \nu_{i,m}\middle| \nu_{i,m} \sim \mathcal{U}(\gT), i=1..N, m=1..M\right\}$
        \Comment Sample parameters of Q augmentations}
        \State $  J_Q(\theta) = \frac{1}{N} \sum_{i=1}^{N} (Q_{\theta}(s_i, a_i) - y_i)^2 $ \mbox{ or } \textcolor{q_aug}{$J_Q(\theta) =\frac{1}{NM} {\sum_{i,m=1}^{N,M}} (Q_{\theta}($}\textcolor{f_func}{$f$}\textcolor{q_aug}{$(s_i, \nu_{i,m}), a_i) - y_i)^2$}
        \State $ \theta \leftarrow \theta - \lambda_{\theta} \nabla_\theta  J_Q(\theta)$ 
        \Comment Update the critic
        \State $\theta' \leftarrow (1 - \tau) \theta' + \tau \theta$
        \Comment Update the critic target
    \EndProcedure
\end{algorithmic}

\label{alg:final}
\end{algorithm*}

\section{Experiments}
\label{section:experiments}
In this section we evaluate our algorithm (\textbf{DrQ}) on the two commonly used benchmarks based on the DeepMind control suite \cite{tassa2018dmcontrol}, namely the PlaNet~\cite{hafner2018planet} and Dreamer~\cite{hafner2019dream} setups. Throughout these experiments all  hyper-parameters of the algorithm are kept fixed: the actor and critic neural networks are trained using the Adam optimizer~\cite{kingma2014adam} with default parameters and a mini-batch size of $512$. For SAC, the soft target update rate $\tau$ is $0.01$, initial temperature is $0.1$, and target network and the actor updates are made  every $2$ critic updates (as in~\cite{yarats2019improving}). We use the image encoder architecture from SAC-AE~\cite{yarats2019improving} and follow their training procedure. The full set of parameters is in~\cref{section:hyperparams}.

\begin{figure}[t!]
    \centering
    %\vspace{-20pt}
    \includegraphics[width=1.0\linewidth]{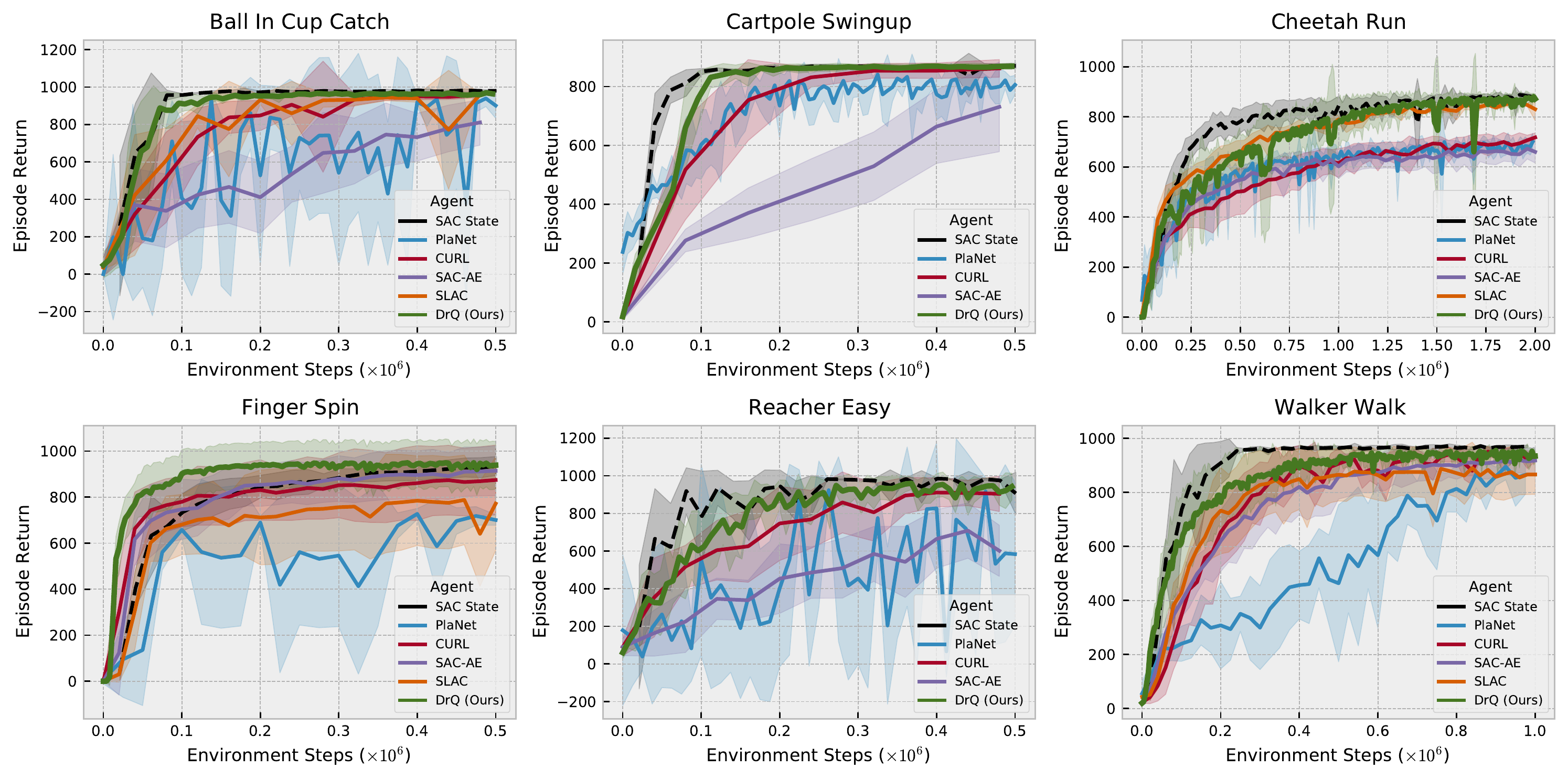}
    \caption{The PlaNet benchmark. Our algorithm (\textbf{DrQ} \textsc{[K=2,M=2]}) outperforms the other methods and demonstrates the state-of-the-art performance. Furthermore, on several tasks \drq is able to  match the upper-bound performance of SAC trained directly on internal state, rather than images. Finally, our algorithm not only shows improved sample-efficiency relative to other approaches, but is also faster in terms of wall clock time.}
    \label{fig:planet_bench}
\end{figure}

\begin{table*}[ht!]
%\vspace{-20pt}

\begin{center}
\begin{small}
\begin{tabular}{|l|ccccc|c|}
\hline
\emph{500k step scores} & DrQ  (Ours) &  CURL & PlaNet & SAC-AE & SLAC &  SAC State \\
\hline
Finger Spin &\textbf{938$\pm$103}&874$\pm$151 &718$\pm$40 &914$\pm$107 &771$\pm$203 &927$\pm$43 \\
Cartpole Swingup &\textbf{868$\pm$10}& 861$\pm$30  &787$\pm$46 &730$\pm$152 & -  &870$\pm$7 \\
Reacher Easy &\textbf{942$\pm$71}& 904$\pm$94  &588$\pm$471 &601$\pm$135 & -  &975$\pm$5 \\
Cheetah Run &\textbf{660$\pm$96}& 500$\pm$91 &568$\pm$21 &544$\pm$50 &629$\pm$74 &772$\pm$60 \\
Walker Walk &\textbf{921$\pm$45}& 906$\pm$56&478$\pm$164 &858$\pm$82 &865$\pm$97 &964$\pm$8 \\
Ball In Cup Catch &\textbf{963$\pm$9} &958$\pm$13 &939$\pm$43 &810$\pm$121 &959$\pm$4 &979$\pm$6 \\
\hline
\emph{100k step scores} & &   &  &  &  &  \\
\hline
Finger Spin &\textbf{901$\pm$104}& 779$\pm$108 &560$\pm$77 &747$\pm$130 &680$\pm$130 &672$\pm$76 \\
Cartpole Swingup &\textbf{759$\pm$92}& 592$\pm$170 &563$\pm$73 &276$\pm$38 & -  &812$\pm$45 \\
Reacher Easy &\textbf{601$\pm$213}& 517$\pm$113 &82$\pm$174 &225$\pm$164 & -  &919$\pm$123 \\
Cheetah Run &344$\pm$67&     307$\pm$48 &165$\pm$123 &252$\pm$173 &\textbf{391$\pm$47}$^*$ &228$\pm$95 \\
Walker Walk &\textbf{612$\pm$164}&  344$\pm$132  &221$\pm$43 &395$\pm$58 &428$\pm$74 &604$\pm$317 \\
Ball In Cup Catch &\textbf{913$\pm$53}& 772$\pm$241 &710$\pm$217 &338$\pm$196 &607$\pm$173 &957$\pm$26 \\
 \hline
\end{tabular}
\end{small}
\end{center}
%\vskip -0.1in
\caption{The PlaNet benchmark at $100$k and $500$k environment steps. Our method (\textbf{DrQ} \textsc{[K=2,M=2]}) outperforms other approaches in both the data-efficient ($100$k) and asymptotic performance ($500$k) regimes. $^*$: SLAC uses $100$k exploration steps which are not counted in the reported values. By contrast, \drq only uses $1000$ exploration steps which are included in the overall step count. }
\label{table:planet_bench}
\end{table*}

Following \cite{henderson2017rlmatters}, the models are trained using $10$ different seeds; for every seed the mean episode returns are computed every $10000$ environment steps, averaging over $10$ episodes. All figures plot the mean performance over the $10$ seeds, together with $\pm$ 1 standard deviation shading. We compare our  \drq approach to leading model-free and model-based approaches: PlaNet~\cite{hafner2018planet}, SAC-AE~\cite{yarats2019improving}, SLAC~\cite{lee2019slac}, CURL~\cite{srinivas2020curl} and Dreamer~\cite{hafner2019dream}. The comparisons use the results provided by the authors of the corresponding papers.

\subsection{DeepMind Control Suite Experiments}
\textbf{PlaNet Benchmark}~\cite{hafner2018planet} consists of six challenging control tasks from \cite{tassa2018dmcontrol} with different traits. The benchmark specifies a different action-repeat hyper-parameter for each of the six tasks\footnote{This means the number of training observations is a fraction of the environment steps (e.g. an episode of $1000$ steps with action-repeat $4$ results in $250$ training observations).}. Following common practice~\cite{hafner2018planet,lee2019slac,yarats2019improving,mnih2013dqn}, we report the performance using true environment steps, thus are invariant to the action-repeat hyper-parameter.
Aside from action-repeat, all other hyper-parameters of our algorithm are fixed across the six tasks, using the values previously detailed.

\cref{fig:planet_bench} compares \drq \textsc{[K=2,M=2]} to PlaNet~\cite{hafner2018planet}, SAC-AE~\cite{yarats2019improving}, CURL~\cite{srinivas2020curl}, SLAC~\cite{lee2019slac}, and an upper bound performance provided by SAC~\cite{haarnoja2018sac} that directly learns from internal states. We use the version of SLAC that performs one gradient update per an environment step to ensure a fair comparison to other approaches. 
\drq achieves state-of-the-art performance on this benchmark on all the tasks, despite being much simpler than other methods. Furthermore, since \drq does not learn a model~\cite{hafner2018planet,lee2019slac} or any auxiliary tasks~\cite{srinivas2020curl}, the wall clock time also compares favorably to the other methods. In~\cref{table:planet_bench} we also compare performance given at a fixed number of environment interactions (e.g. $100$k and $500$k). Furthermore, in~\cref{section:robust_study_full} we demonstrate that \drq is robust to significant changes in hyper-parameter settings.

\textbf{Dreamer Benchmark} is a more extensive testbed that was introduced in Dreamer~\cite{hafner2019dream}, featuring a diverse set of tasks from the DeepMind control suite. 
Tasks involving sparse reward were excluded (e.g. Acrobot and Quadruped) since they require modification of SAC to incorporate multi-step returns \cite{barth-maron2018d4pg}, which is beyond the scope of this work. We evaluate on the remaining $15$ tasks, fixing the action-repeat hyper-parameter to $2$, as in Dreamer~\cite{hafner2019dream}.

\begin{figure}[ht!]
    \centering
    
    \includegraphics[width=1.\linewidth]{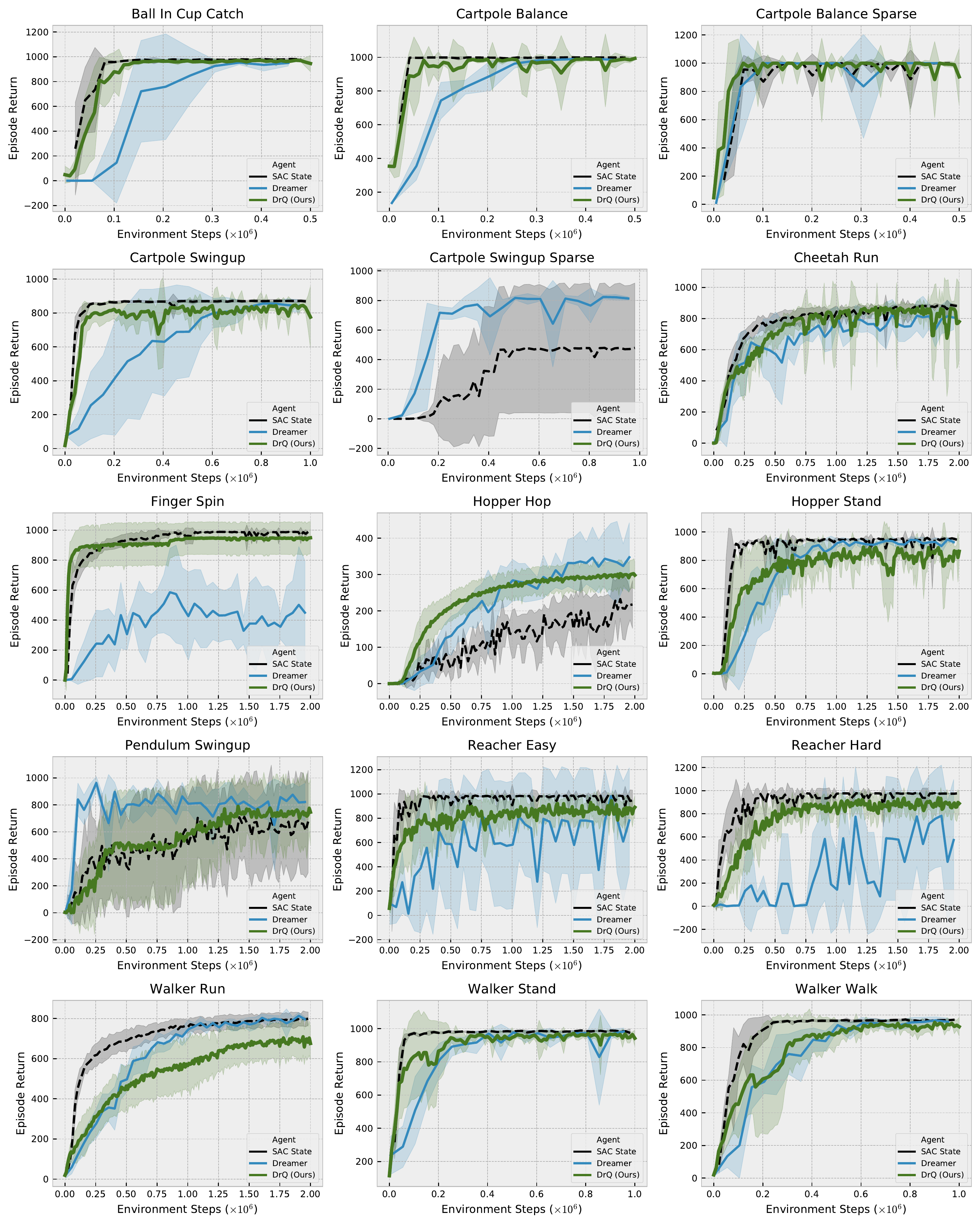}
    \caption{ The Dreamer benchmark. Our method (\drq \textsc{[K=2,M=2]}) again demonstrates superior performance over Dreamer on $12$ out $15$ selected tasks. In many cases it also reaches the upper-bound performance of SAC that learns directly from states.}
    \label{fig:dreamer_bench}
\end{figure}

We compare \drq \textsc{[K=2,M=2]} to Dreamer~\cite{hafner2019dream} and the upper-bound performance of SAC~\cite{haarnoja2018sac} from states\footnote{No other publicly reported results are available for the other methods due to the recency of the Dreamer~\cite{hafner2019dream} benchmark.}. Again, we keep all the hyper-parameters of our algorithm fixed across all the tasks. In~\cref{fig:dreamer_bench}, \drq demonstrates the state-of-the-art results by collectively outperforming Dreamer~\cite{hafner2019dream}, although Dreamer is superior on $3$ of the $15$ tasks (Walker Run, Cartpole Swingup Sparse and Pendulum Swingup). On many tasks \drq approaches the upper-bound performance of SAC~\cite{haarnoja2018sac} trained directly on states.  

\subsection{Atari 100k Experiments}
\begin{figure}[t!]
     \centering
     %\vspace{-20pt}
     \includegraphics[width=1\linewidth]{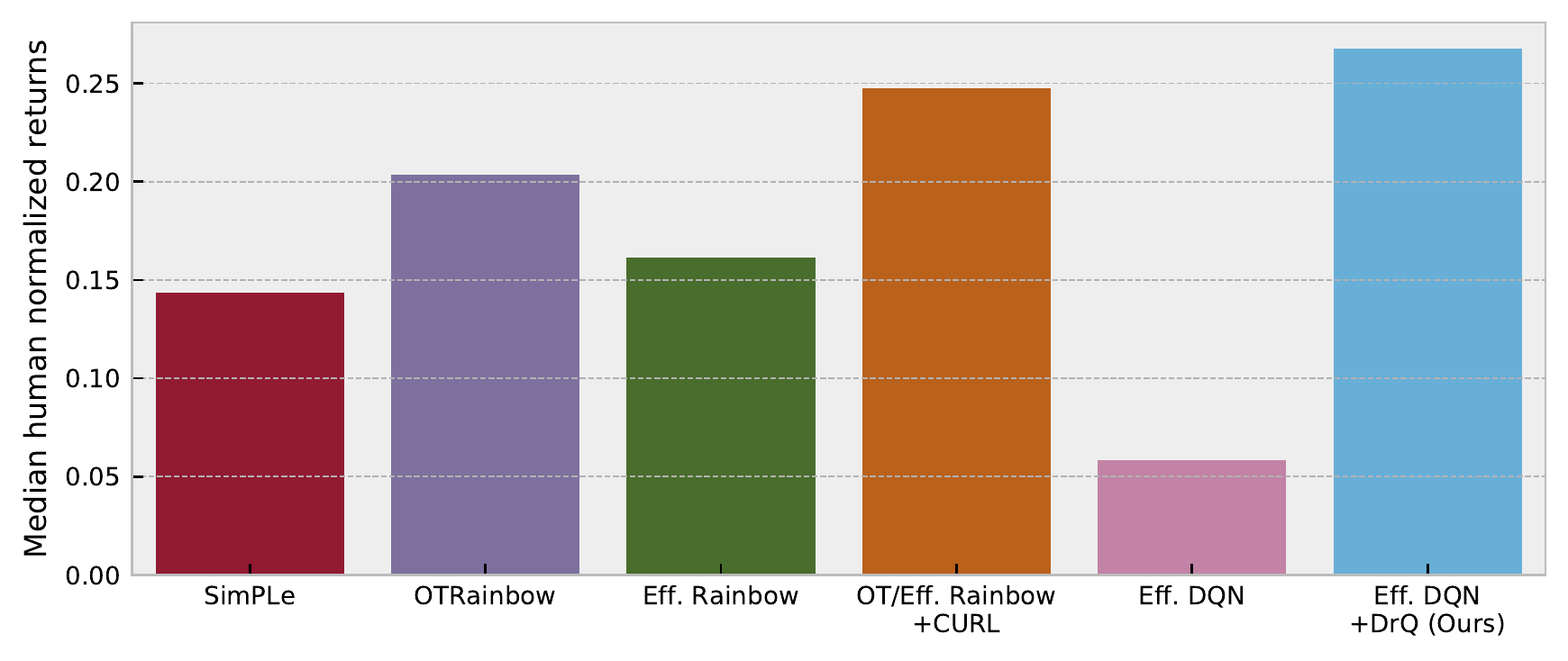}
     \caption{The Atari 100k benchmark. Compared to a set of leading baselines, our method (\drq \textsc{[K=1,M=1]}, combined with Efficient DQN) achieves the state-of-the-art performance, despite being considerably simpler. Note the large improvement that results from adding \drq to Efficient DQN (pink vs cyan). By contrast, the gains from CURL, that utilizes tricks from both Data Efficient Rainbow and OTRainbow, are more modest over the underlying RL methods.}
     \label{fig:atari_bench}
\end{figure}

We evaluate \drq \textsc{[K=1,M=1]} on the recently introduced Atari 100k~\cite{kaiser2018simple} benchmark -- a sample-constrained evaluation for discrete control algorithms. The underlying RL approach to which \drq is applied is a DQN, combined with double Q-learning~\cite{hasselt2015doubledqn}, n-step returns~\cite{mnih2016a3c}, and dueling critic architecture~\cite{wang2015dueling}. As per common practice~\cite{kaiser2018simple,hasselt2019derainbow}, we evaluate our agent for 125k environment steps at the end of training and average its performance over $5$ random seeds.~\cref{fig:atari_bench} shows the median human-normalized episode returns performance (as in~\cite{mnih2013dqn}) of the underlying model, which we refer to as Efficient DQN, in pink. When \drq is added there is a significant increase in performance (cyan), surpassing OTRainbow~\cite{kielak2020do} and Data Efficient Rainbow~\cite{hasselt2019derainbow}. \drq is also  superior to CURL \cite{srinivas2020curl} that uses an auxiliary loss built on top of a hybrid between OTRainbow and Efficient rainbow. \drq combined with Efficient DQN thus achieves state-of-the-art performance, despite being significantly simpler than the other approaches. The experimental setup is detailed in~\cref{section:atari_hyperparams} and full results can be found in~\cref{section:full_atari_results}. 

%% Save for rebuttal -- doesn't really add much....
%Moreover, there is no direct comparison between CURL and this version of Rainbow making it difficult to understand the importance of contrastive loss. 

\section{Related Work}

\textbf{Computer Vision} Data augmentation via image transformations has been used to improve generalization since the inception of convolutional networks \cite{Becker92,simard2003best,lecun1989backpropagation,cirecsan2011high,ciregan2012multi}. Following AlexNet \cite{krizhevsky2012imagenet}, they have become a standard part of training pipelines. For object classification tasks, the transformations are selected to avoid changing the semantic category, i.e. translations, scales, color shifts, etc. Perturbed versions of input examples are used to expand the training set and no adjustment to the training algorithm is needed. While a similar set of transformations are potentially applicable to control tasks, the RL context does require modifications to be made to the underlying algorithm. 

Data augmentation methods have also been used in the context of self-supervised learning. \cite{Dosovitsky16} use per-exemplar perturbations in a unsupervised classification framework. More recently, a several approaches \cite{chen2020simple,  he2019moco, misra2019self, henaff2019cpcv2} have used invariance to imposed image transformations in contrastive learning schemes, producing state-of-the-art results on downstream recognition tasks. By contrast, our scheme addresses control tasks, utilizing different types of invariance.

\textbf{Regularization in RL} Some early attempts to learn RL function approximators used $\ell_2$ regularization of the Q~\cite{Farahmand08, Yan17} function. Another approach is entropy regularization~\cite{ziebart2008maxent, haarnoja2018sac, nachum2017bridging, williams1991function}, where causal entropy is added to the rewards, making the Q-function smoother and facilitating optimization~\cite{ahmed2018understanding}. Prior work has explored regularization of the neural network approximator in deep RL, e.g. using dropout~\cite{Farebrother18} and cutout~\cite{cobbe2018quantifying} techniques. See \cite{liu2019regularization} for a comprehensive evaluation of different network regularization methods. In contrast, our approach directly regularizes the Q-function in a data-driven way that incorporates knowledge of task invariances, as opposed to generic priors.

\textbf{Generalization between Tasks and Domains} A range of datasets have been introduced with the explicit aim of improving generalization in RL through deliberate variation of the scene colors/textures/backgrounds/viewpoints. These include Robot Learning in Homes~\cite{gupta2018robot}, Meta-World \cite{yu2019meta}, the ProcGen benchmark~\cite{ProcGen}. There are also domain randomization techniques \cite{tobin2017domain, slaoui2019robust} which synthetically apply similar variations, but assume control of the data generation procedure, in contrast to our method. Furthermore, these works address generalization between domains (e.g. synthetic-to-real or different game levels), whereas our work focuses on a single domain and task. In concurrent work, RAD~\cite{laskin2020rad} also demonstrates that image augmentation can improve sample efficiency and generalization of RL algorithms. However, RAD represents a specific instantiation of our algorithm when $\textsc{[K=1,M=1]}$ and different image augmentations are used.

\textbf{Continuous Control from Pixels} There are a variety of methods addressing the sample-efficiency of RL algorithms that directly learn from pixels. The most prominent approaches for this can be classified into two groups, model-based and model-free methods. The model-based methods attempt to learn the system dynamics in order to acquire a compact latent representation of high-dimensional observations to later perform policy search~\cite{hafner2018planet,lee2019slac,hafner2019dream}. In contrast, the model-free methods either learn the latent representation indirectly by optimizing the RL objective~\cite{barth-maron2018d4pg,abdolmaleki2018maximum} or by employing auxiliary losses that provide additional supervision~\cite{yarats2019improving,srinivas2020curl,sermanet2018time,dwibedi2018visrepr}. Our approach is complementary to these methods and can be combined with them to improve performance.

\section{Conclusion}

We have introduced a simple regularization technique that significantly improves the performance of SAC trained directly from image pixels on standard continuous control tasks. Our method is easy to implement and adds a negligible computational burden. We compared our method to state-of-the-art approaches on both DeepMind control suite, where we demonstrated that it outperforms them on the majority of tasks, and Atari 100k benchmarks, where it outperforms other methods in the median metric. Furthermore, we demonstrate the method to be robust to the choice of hyper-parameters.

%Authors are required to include a statement of the broader impact of their work, including its ethical aspects and future societal consequences. 
%Authors should discuss both positive and negative outcomes, if any. For instance, authors should discuss a) 
%who may benefit from this research, b) who may be put at disadvantage from this research, c) what are the consequences of failure of the system, and d) whether the task/method leverages
%biases in the data. If authors believe this is not applicable to them, authors can simply state this.

%Use unnumbered first level headings for this section, which should go at the end of the paper. {\bf Note that this section does not count towards the eight pages of content that are allowed.}

\section{Acknowledgements}
We would like to thank Danijar Hafner, Alex Lee, and Michael Laskin for sharing performance data for the Dreamer~\cite{hafner2019dream} and PlaNet~\cite{hafner2018planet}, SLAC~\cite{lee2019slac}, and CURL~\cite{srinivas2020curl} baselines respectively. Furthermore, we would like to thank Roberta Raileanu for helping with the architecture experiments. Finally, we would like to thank Ankesh Anand for helping us finding an error in our evaluation script for the Atari 100k benchmark experiments.

\bibliography{main}
\bibliographystyle{plain}

\includeappendixtrue % comment it out to hide appendix

\ifincludeappendix

\newpage
\appendix
\section*{Appendix}

\section{Extended Background}
\label{section:extended_background}

\paragraph{Reinforcement Learning from Images} We formulate image-based control as an infinite-horizon partially observable Markov decision process (POMDP) \cite{bellman1957mdp,kaelbling1998planning}. An POMDP can be described as the tuple $(\gO, \gA, p, r, \gamma)$, where $\gO$ is the high-dimensional observation space (image pixels), $\gA$ is the action space, the transition dynamics $p = Pr(o'_{t}|o_{\leq t},a_{t})$ capture the probability distribution over the next observation $o'_t$ given the history of previous observations $o_{\leq t}$ and current action $a_{t}$, $r: \gO \times \gA \rightarrow \R$ is the reward function that maps the current observation and action to a reward $r_t  = r(o_{\leq t}, a_t)$, and $\gamma \in [0, 1)$ is a discount factor. Per common practice~\cite{mnih2013dqn}, throughout the paper the POMDP is converted into an MDP~\cite{bellman1957mdp} by stacking several consecutive image observations into a state $s_t = \{o_t, o_{t-1}, o_{t-2},\ldots\}$. For simplicity we redefine  the transition dynamics $p = Pr(s'_{t}|s_t,a_{t})$ and the reward function $r_t = r(s_t, a_t)$. We then aim to find a policy $\pi(a_t|s_t)$ that maximizes the cumulative discounted return $\E_{\pi}[\sum_{t=1}^{\infty} \gamma^t r_t | a_t \sim \pi(\cdot|s_t), s'_t \sim p(\cdot|s_t, a_t), s_1 \sim p(\cdot)]$.

\paragraph{Soft Actor-Critic} 
The Soft Actor-Critic (SAC)~\cite{haarnoja2018sac} learns a state-action value function $Q_\theta$, a stochastic policy $\pi_\theta$  and a temperature $\alpha$ to find an optimal policy for an MDP $(\gS, \gA, p, r, \gamma)$ by optimizing a $\gamma$-discounted maximum-entropy objective~\cite{ziebart2008maxent}.
$\theta$ is used generically to denote the parameters updated through training in each part of the model. The actor policy $\pi_\theta(a_t|s_t)$ is a parametric $\mathrm{tanh}$-Gaussian that given $s_t$ samples  $a_t = \mathrm{tanh}(\mu_\theta(s_t)+ \sigma_\theta(s_t) \epsilon)$, where $\epsilon \sim \gN(0, 1)$ and $\mu_\theta$ and $\sigma_\theta$ are parametric mean and standard deviation.

The policy evaluation step  learns  the critic $Q_\theta(s_t, a_t)$ network by optimizing a single-step of the soft Bellman residual
\begin{align*}
    J_Q(\gD) &= \E_{\substack{( s_t,a_t, s'_t) \sim \gD \\ a_t' \sim \pi(\cdot|s_t')}}[(Q_\theta(s_t, a_t) - y_t)^2] \\
    y_t &= r(s_t, a_t) + \gamma [Q_{\theta'}(s'_t, a'_t) - \alpha \log \pi_\theta(a'_t|s'_t)] , \\
\end{align*}
where $\gD$ is a replay buffer of transitions, $\theta'$ is an exponential moving average of the weights as done in~\cite{lillicrap2015ddpg}. SAC uses clipped double-Q learning~\cite{hasselt2015doubledqn,fujimoto2018td3}, which we omit from our notation for simplicity but employ in practice.

The policy improvement step then fits the actor policy $\pi_\theta(a_t|s_t)$ network by optimizing the objective
\begin{align*}
    J_\pi(\gD) &= \E_{s_t \sim \gD}[ \KL(\pi_\theta(\cdot|s_t) || \exp\{\frac{1}{\alpha}Q_\theta(s_t, \cdot)\})].
\end{align*}
Finally, the temperature $\alpha$ is learned with the loss
\begin{align*}
    J_\alpha(\gD) &= \E_{\substack{s_t \sim \gD \\ a_t \sim \pi_\theta(\cdot|s_t)}}[-\alpha \log \pi_\theta(a_t|s_t) - \alpha \bar\gH],
\end{align*}
where $\bar\gH \in \R$ is the target entropy hyper-parameter that the policy tries to match, which in practice is usually set to  $\bar\gH=-|\gA|$.
\paragraph{Deep Q-learning} DQN~\cite{mnih2013dqn} also learns a convolutional neural net to approximate Q-function over states and actions. The main difference is that DQN operates on discrete actions spaces, thus the policy can be directly inferred from Q-values. The parameters of DQN are updated by optimizing the squared residual error

\begin{align*}
    J_Q(\gD) &= \E_{( s_t,a_t, s'_t) \sim \gD}[(Q_\theta(s_t, a_t) - y_t)^2] \\
    y_t &= r(s_t, a_t) + \gamma \max_{a'} Q_{\theta'}(s'_t, a') . \\
\end{align*}

In practice, the standard version of DQN is frequently combined with a set of tricks that improve performance and training stability, wildly known as Rainbow~\cite{hasselt2015doubledqn}.

\section{The DeepMind Control Suite Experiments Setup}
\label{section:hyperparams}
Our PyTorch SAC~\cite{haarnoja2018sac} implementation is based off of~\cite{yarats2020pytorch_sac}.
\subsection{Actor and Critic Networks}
We employ clipped double Q-learning~\cite{hasselt2015doubledqn, fujimoto2018td3} for the critic, where each $Q$-function is parametrized as a 3-layer MLP with \texttt{ReLU} activations after each layer except of the last. The actor is also a 3-layer MLP with \texttt{ReLU}s that outputs mean and covariance for the diagonal Gaussian that represents the policy. The hidden dimension is set to $1024$ for both the critic and actor.

\subsection{Encoder Network} 
We employ an encoder architecture from~\cite{yarats2019improving}. This encoder consists of four convolutional layers with $3\times 3$ kernels and $32$ channels. The \texttt{ReLU} activation is applied after each conv layer. We use stride to $1$ everywhere, except of the first conv layer, which has stride $2$. The output of the convnet is feed into a single fully-connected layer normalized by \texttt{LayerNorm}~\cite{ba2016layernorm}. Finally, we apply \texttt{tanh} nonlinearity to the $50$ dimensional output of the fully-connected layer. We initialize the weight matrix of fully-connected and convolutional layers with the orthogonal initialization~\cite{saxe2013ortho} and set the bias to be zero.

The actor and critic networks both have separate encoders, although we share the weights of the conv layers between them. Furthermore, only the critic optimizer is allowed to update these weights (e.g. we stop the gradients from the actor before they propagate to the shared conv layers).

\subsection{Training and Evaluation Setup}

Our agent first collects $1000$ seed observations using a random policy. The further training observations are collected by sampling actions from the current policy. We perform one training update every time we receive a new observation. In cases where we use action repeat, the number of training observations is only a fraction of the environment steps (e.g. a $1000$ steps episode at action repeat $4$ will only results into $250$ training observations). We evaluate our agent every $10000$ true environment steps by computing the average episode return over $10$ evaluation episodes. During evaluation we take the mean policy action instead of sampling.
\subsection{PlaNet and Dreamer Benchmarks}
\label{section:hyperparams:training_setup}

We consider two evaluation setups that were introduced in PlaNet~\cite{hafner2018planet} and Dreamer~\cite{hafner2019dream}, both using tasks from the DeepMind control suite~\cite{tassa2018dmcontrol}. The PlaNet benchmark consists of six tasks of various traits. Importantly, the benchmark proposed to use a different action repeat hyper-parameter for each task, which we summarize in~\cref{table:action_repeat}.

The Dreamer benchmark considers an extended set of tasks, which makes it more difficult that the PlaNet setup. Additionally, this benchmark requires to use the same set hyper-parameters for each task, including action repeat (set to $2$), which further increases the difficulty. 

\begin{table}[ht!]
\centering
\begin{tabular}{|l|c|}
\hline
Task name        & Action repeat \\
\hline
Cartpole Swingup &  $8$ \\
Reacher Easy &  $4$ \\
Cheetah Run & $4$ \\
Finger Spin & $2$ \\
Ball In Cup Catch & $4$ \\
Walker Walk & $2$ \\
\hline
\end{tabular}
\caption{\label{table:action_repeat} The action repeat hyper-parameter used for each task in the PlaNet benchmark.}
\end{table}

\subsection{Pixels Preprocessing}
We construct an observational input as an $3$-stack of consecutive frames~\cite{mnih2013dqn}, where each frame is a RGB rendering of size $ 84 \times 84$ from the $0$th camera. We then divide each pixel by $255$ to scale it down to $[0, 1]$ range.

\subsection{Other Hyper Parameters}
Due to computational constraints for all the continuous control ablation experiments in the main paper and appendix we use a minibatch size of $128$, while for the main results we use minibatch of size $512$. In~\cref{table:hyper_params} we provide a comprehensive overview of all the other hyper-parameters.

\begin{table}[hb!]
\centering
\begin{tabular}{|l|c|}
\hline
Parameter        & Setting \\
\hline
Replay buffer capacity & $100000$ \\
Seed steps & $1000$ \\
Ablations minibatch size & $128$ \\
Main results minibatch size & $512$ \\
Discount $\gamma$ & $0.99$ \\
Optimizer & Adam \\
Learning rate & $10^{-3}$ \\
Critic target update frequency & $2$ \\
Critic Q-function soft-update rate $\tau$ & $0.01$ \\
Actor update frequency & $2$ \\
Actor log stddev bounds & $[-10, 2]$ \\
Init temperature & $0.1$ \\

\hline
\end{tabular}\\
\caption{\label{table:hyper_params} An overview of used hyper-parameters in the DeepMind control suite experiments.}
\end{table}

\newpage

\section{The Atari 100k Experiments Setup}
For ease of reproducibility in~\cref{table:atari_hyper_params} we report the hyper-parameter settings used in the Atari 100k experiments. We largely reuse the hyper-parameters from OTRainbow~\cite{kielak2020do}, but adapt them for DQN~\cite{mnih2013dqn}. Per common practise, we average performance of our agent over $5$ random seeds. The evaluation is done for 125k environment steps at the end of training for 100k environment steps.
\label{section:atari_hyperparams}

\begin{table}[hb!]
\centering
\begin{tabular}{|l|c|}
\hline
Parameter         & Setting \\
\hline
Data augmentation & Random shifts and Intensity \\
Grey-scaling & True \\
Observation down-sampling & $84 \times 84$ \\
Frames stacked & $4$ \\
Action repetitions & $4$ \\
Reward clipping & $[-1, 1]$ \\
Terminal on loss of life  & True \\
Max frames per episode & $108$k \\
Update  &  Double Q \\
Dueling & True \\
Target network: update period & $1$ \\
Discount factor &  $0.99$ \\
Minibatch size  & $32$ \\
Optimizer  & Adam \\
Optimizer: learning rate & $0.0001$\\
Optimizer: $\beta_1$  & $0.9$ \\
Optimizer: $\beta_2$ &  $0.999$ \\
Optimizer: $\epsilon$ &  $0.00015$ \\
Max gradient norm & $10$ \\
Training steps &  $100$k \\
Evaluation steps & $125$k \\
Min replay size for sampling& $1600$\\
Memory size & Unbounded \\
Replay period every & $1$ step \\
Multi-step return length & $10$ \\
Q network: channels  & $32, 64, 64$ \\ 
Q network: filter size  & $8 \times 8$, $4 \times 4$, $3 \times 3$ \\
Q network: stride  & $4, 2, 1$ \\
Q network: hidden units  & $512$ \\
Non-linearity & \texttt{ReLU}\\
Exploration & $\epsilon$-greedy \\
$\epsilon$-decay & $5000$ \\
\hline
\end{tabular}\\
\caption{\label{table:atari_hyper_params} A complete overview of hyper parameters used in the Atari 100k experiments.}
\end{table}

\newpage

\section{Full Atari 100k Results}
\label{section:full_atari_results}
Besides reporting in~\cref{fig:atari_bench} median human-normalized episode returns over  the $26$ Atari games used in~\cite{kaiser2018simple}, we also provide the mean episode return for each individual game in~\cref{table:atari_bench}.  Human/Random scores are taken from~\cite{hasselt2019derainbow} to be consistent with the established setup.

\begin{table*}[ht!]
\vskip 0.15in
\begin{center}
\begin{small}
\scalebox{.73}{
\begin{tabular}{|l|cc|cccccc|}
\hline
\multirow{2}{*}{Game} & \multirow{2}{*}{Human} & \multirow{2}{*}{Random}&   \multirow{2}{*}{SimPLe} & \multirow{2}{*}{OTRainbow} & \multirow{2}{*}{Eff. Rainbow} & OT/Eff. Rainbow & \multirow{2}{*}{Eff. DQN} & Eff. DQN  \\
  &  & & & & & +CURL & & +DrQ (Ours) \\
\hline
Alien & 7127.7 & 227.8 & 616.9 & 824.7 & 739.9 & \textbf{1148.2} & 558.1 & 771.2\\
Amidar & 1719.5 & 5.8 & 88.0 & 82.8 & 188.6 & \textbf{232.3} & 63.7 & 102.8\\
Assault & 742.0 & 222.4 & 527.2 & 351.9 & 431.2 & 543.7 & \textbf{589.5} & 452.4\\
Asterix & 8503.3 & 210.0 & \textbf{1128.3} & 628.5 & 470.8 & 524.3 & 341.9 & 603.5\\
BankHeist & 753.1 & 14.2 & 34.2 & 182.1 & 51.0 & \textbf{193.7} & 74.0 & 168.9\\
BattleZone & 37187.5 & 2360.0 & 5184.4 & 4060.6 & 10124.6 & 11208.0 & 4760.8 & \textbf{12954.0}\\
Boxing & 12.1 & 0.1 & \textbf{9.1} & 2.5 & 0.2 & 4.8 & -1.8 & 6.0\\
Breakout & 30.5 & 1.7 & 16.4 & 9.8 & 1.9 & \textbf{18.2} & 7.3 & 16.1\\
ChopperCommand & 7387.8 & 811.0 & \textbf{1246.9} & 1033.3 & 861.8 & 1198.0 & 624.4 & 780.3\\
CrazyClimber & 35829.4 & 10780.5 & \textbf{62583.6} & 21327.8 & 16185.3 & 27805.6 & 5430.6 & 20516.5\\
DemonAttack & 1971.0 & 152.1 & 208.1 & 711.8 & 508.0 & 834.0 & 403.5 & \textbf{1113.4}\\
Freeway & 29.6 & 0.0 & 20.3 & 25.0 & \textbf{27.9} & \textbf{27.9} & 3.7 & 9.8\\
Frostbite & 4334.7 & 65.2 & 254.7 & 231.6 & 866.8 & \textbf{924.0} & 202.9 & 331.1\\
Gopher & 2412.5 & 257.6 & 771.0 & 778.0 & 349.5 & \textbf{801.4} & 320.8 & 636.3\\
Hero & 30826.4 & 1027.0 & 2656.6 & 6458.8 & \textbf{6857.0} & 6235.1 & 2200.1 & 3736.3\\
Jamesbond & 302.8 & 29.0 & 125.3 & 112.3 & 301.6 & \textbf{400.1} & 133.2 & 236.0\\
Kangaroo & 3035.0 & 52.0 & 323.1 & 605.4 & 779.3 & 345.3 & 448.6 & \textbf{940.6}\\
Krull & 2665.5 & 1598.0 & \textbf{4539.9} & 3277.9 & 2851.5 & 3833.6 & 2999.0 & 4018.1\\
KungFuMaster & 22736.3 & 258.5 & \textbf{17257.2} & 5722.2 & 14346.1 & 14280.0 & 2020.9 & 9111.0\\
MsPacman & 6951.6 & 307.3 & 1480.0 & 941.9 & 1204.1 & \textbf{1492.8} & 872.0 & 960.5\\
Pong & 14.6 & -20.7 & \textbf{12.8} & 1.3 & -19.3 & 2.1 & -19.4 & -8.5\\
PrivateEye & 69571.3 & 24.9 & 58.3 & 100.0 & 97.8 & 105.2 & \textbf{351.3} & -13.6\\
Qbert & 13455.0 & 163.9 & \textbf{1288.8} & 509.3 & 1152.9 & 1225.6 & 627.5 & 854.4\\
RoadRunner & 7845.0 & 11.5 & 5640.6 & 2696.7 & \textbf{9600.0} & 6786.7 & 1491.9 & 8895.1\\
Seaquest & 42054.7 & 68.4 & \textbf{683.3} & 286.9 & 354.1 & 408.0 & 240.1 & 301.2\\
UpNDown & 11693.2 & 533.4 & \textbf{3350.3} & 2847.6 & 2877.4 & 2735.2 & 2901.7 & 3180.8\\
\hline
Median human-normalised & \multirow{2}{*}{1.000} & \multirow{2}{*}{0.000} & \multirow{2}{*}{0.144} & \multirow{2}{*}{0.204} & \multirow{2}{*}{0.161} & \multirow{2}{*}{0.248} & \multirow{2}{*}{0.058} & \multirow{2}{*}{\textbf{0.268}}\\
episode returns &  & & & & & & & \\
\hline
\end{tabular}}
\end{small}
\end{center}

\caption{ Mean episode returns on each of $26$ Atari games from the setup in~\cite{kaiser2018simple}. The results are recorded at the end of training and averaged across $5$ random seeds (the CURL's results are averaged over $3$ seeds as reported in~\cite{srinivas2020curl}). On each game we mark as bold the highest score. Our method demonstrates better overall performance (as reported in~\cref{fig:atari_bench}).
}
\label{table:atari_bench}
\end{table*}

\section{Image Augmentations Ablation}
\label{section:aug_ablation}

Following~\cite{chen2020simple}, we evaluate popular image augmentation techniques, namely random shifts, cutouts, vertical and horizontal flips, random rotations and imagewise intensity jittering. Below, we provide a comprehensive overview of each augmentation. Furthermore, we examine effectiveness of these techniques in~\cref{fig:aug_ablation}. 

\paragraph{Random Shift} We bring our attention to random shifts that are commonly used to regularize neural networks trained on small images~\cite{Becker92,simard2003best,lecun1989backpropagation,cirecsan2011high,ciregan2012multi}. In our implementation of this method images of size $84 \times 84$ are padded each side by $4$ pixels (by repeating boundary pixels) and then randomly cropped back to the original $84 \times 84 $ size.

\paragraph{Cutout} Cutouts introduced in \cite{devries2017improved} represent a generalization of Dropout~\cite{hinton2012improving}. Instead of masking individual pixels cutouts mask square regions. Since image pixels can be highly correlated, this technique is proven to improve training of neural networks.

\paragraph{Horizontal/Vertical Flip} This technique simply flips an image either horizontally or vertically with probability $0.1$.

\paragraph{Rotate} Here, an image is rotated by $r$ degrees, where $r$ is uniformly sampled from  $[-5, -5]$.

\paragraph{Intensity} Each $N\times C\times84\times84$ image tensor is multiplied by a single scalar $s$, which is computed as $s = \mu + \sigma \cdot \mathrm{clip}(r, -2, 2)$, where $r \sim \gN(0, 1)$. For our experiments we use $\mu=1.0$ and $\sigma=0.1$.

\begin{figure}[ht!]
    \centering
    \includegraphics[width=1.\linewidth]{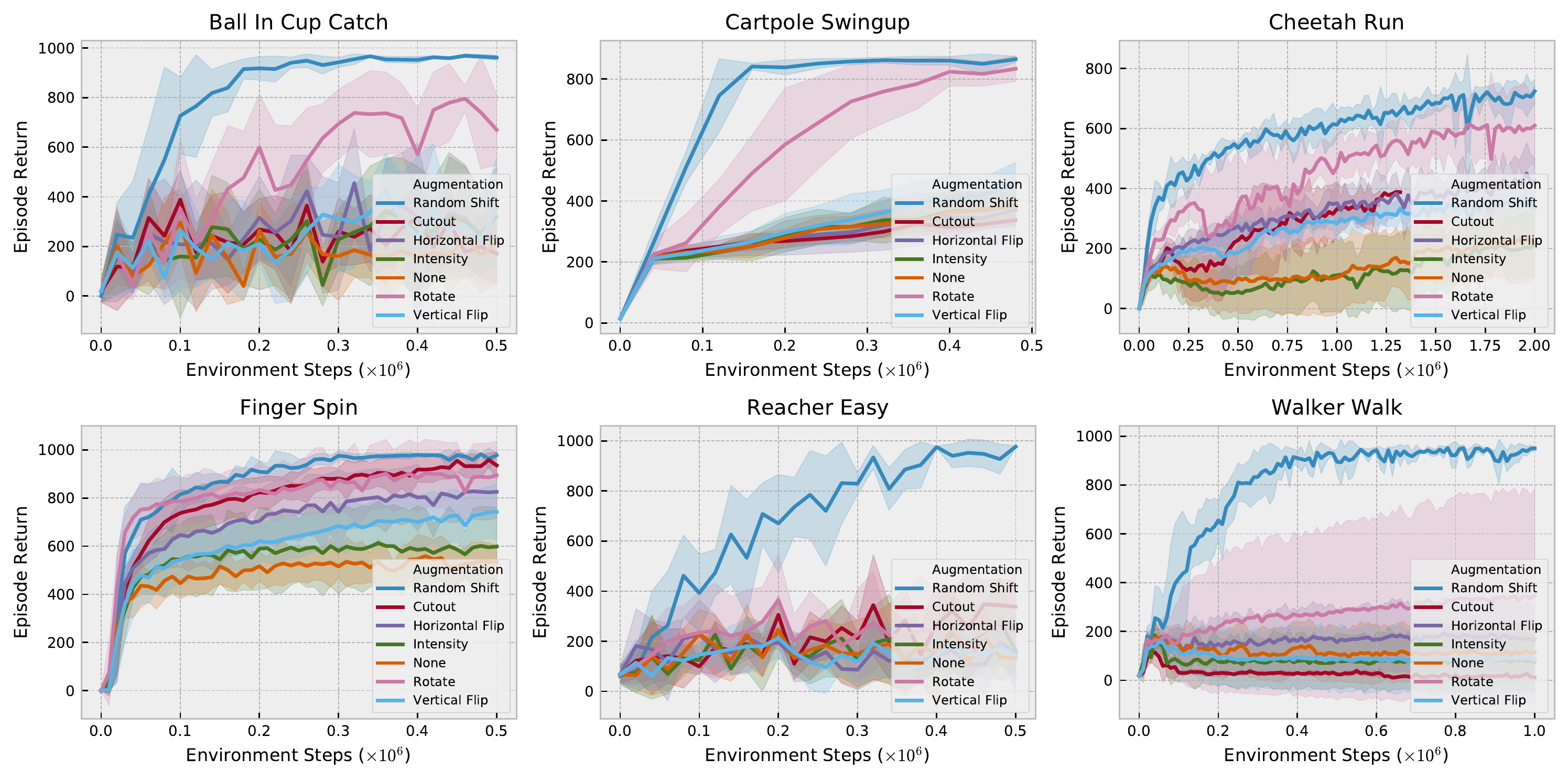}
    \caption{Various image augmentations have different effect on the agent's performance. Overall, we conclude that using image augmentations helps to fight overfitting. Moreover, we notice that random shifts proven to be the most effective technique for tasks from the DeepMind control suite. }
    \label{fig:aug_ablation}
\end{figure}

\paragraph{Implementation}
Finally, we provide Python-like implementation for the aforementioned augmentations powered by Kornia~\cite{riba2020kornia}.

\begin{verbatim}
import torch
import torch.nn as nn
import kornia.augmentation as aug 

random_shift = nn.Sequential(nn.ReplicationPad2d(4),aug.RandomCrop((84, 84)))

cutout = aug.RandomErasing(p=0.5)

h_flip = aug.RandomHorizontalFlip(p=0.1)

v_flip = aug.RandomVerticalFlip(p=0.1)

rotate = aug.RandomRotation(degrees=5.0)

intensity = Intensity(scale=0.05)

class Intensity(nn.Module):
  def __init__(self, scale):
    super().__init__()
    self.scale = scale

  def forward(self, x):
    r = torch.randn((x.size(0), 1, 1, 1), device=x.device)
    noise = 1.0 + (self.scale * r.clamp(-2.0, 2.0))
    return x * noise

\end{verbatim}
\newpage
\section{K and M Hyper-parameters Ablation}
\label{section:km}
We further ablate the \textsc{K,M} hyper-parameters from~\cref{alg:final} to understand their effect on performance. In~\cref{fig:km} we observe that increase values of \textsc{K,M} improves the agent's performance. We choose to use the \textsc{[K=2,M=2]} parametrization as it strikes a good balance between performance and computational demands.

\begin{figure}[ht!]
    \centering
    \includegraphics[width=1.\linewidth]{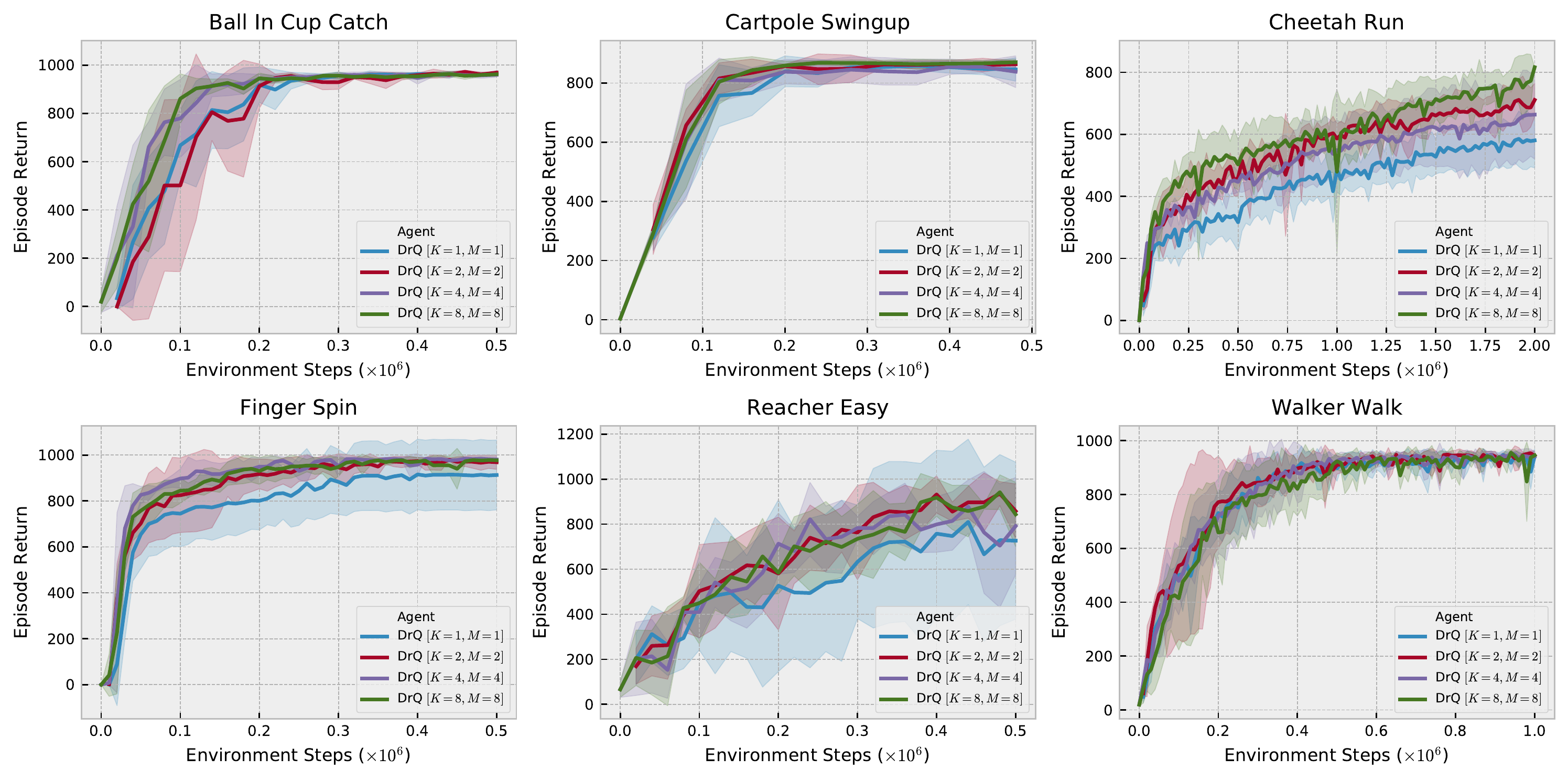}
    \caption{Increasing values of \textsc{K,M} hyper-parameters generally correlates positively with the agent's performance, especially on the harder tasks, such as Cheetah Run. }
    \label{fig:km}
\end{figure}

\section{Robustness Investigation}

To demonstrate the robustness of our approach~\cite{henderson2017rlmatters}, we perform a comprehensive study on the effect  different hyper-parameter choices have on performance. 
A review of prior work ~\cite{hafner2018planet,hafner2019dream,lee2019slac,srinivas2020curl} shows consistent values for discount $\gamma=0.99$ and target update rate $\tau=0.01$ parameters, but variability on network architectures, mini-batch sizes, learning rates. 
Since our method is based on SAC~\cite{haarnoja2018sac}, we also check whether the initial value of the temperature is important, as it plays a crucial role in the initial phase of exploration.  We omit search over network architectures since \cref{fig:encoders_b} shows our method to be robust to the exact choice.
We thus focus on three hyper-parameters: mini-batch size, learning rate, and initial temperature.

Due to computational demands, experiments are restricted to a subset of tasks from \cite{tassa2018dmcontrol}: Walker Walk, Cartpole Swingup, and Finger Spin. These were selected to be diverse, requiring different behaviors including locomotion and goal reaching.
A grid search is performed over mini-batch sizes $\{128, 256, 512\}$, learning rates $\{0.0001, 0.0005, 0.001, 0.005\}$, and initial temperatures $\{0.005, 0.01, 0.05, 0.1\}$.
We follow the experimental setup from~\cref{section:hyperparams}, except that only $3$ seeds are used due to the computation limitations, but since variance is low the results are representative.

\label{section:robust_study_full}
\begin{figure}[H]
    \centering
    \subfloat[Walker Walk.]{\includegraphics[width=1.\linewidth]{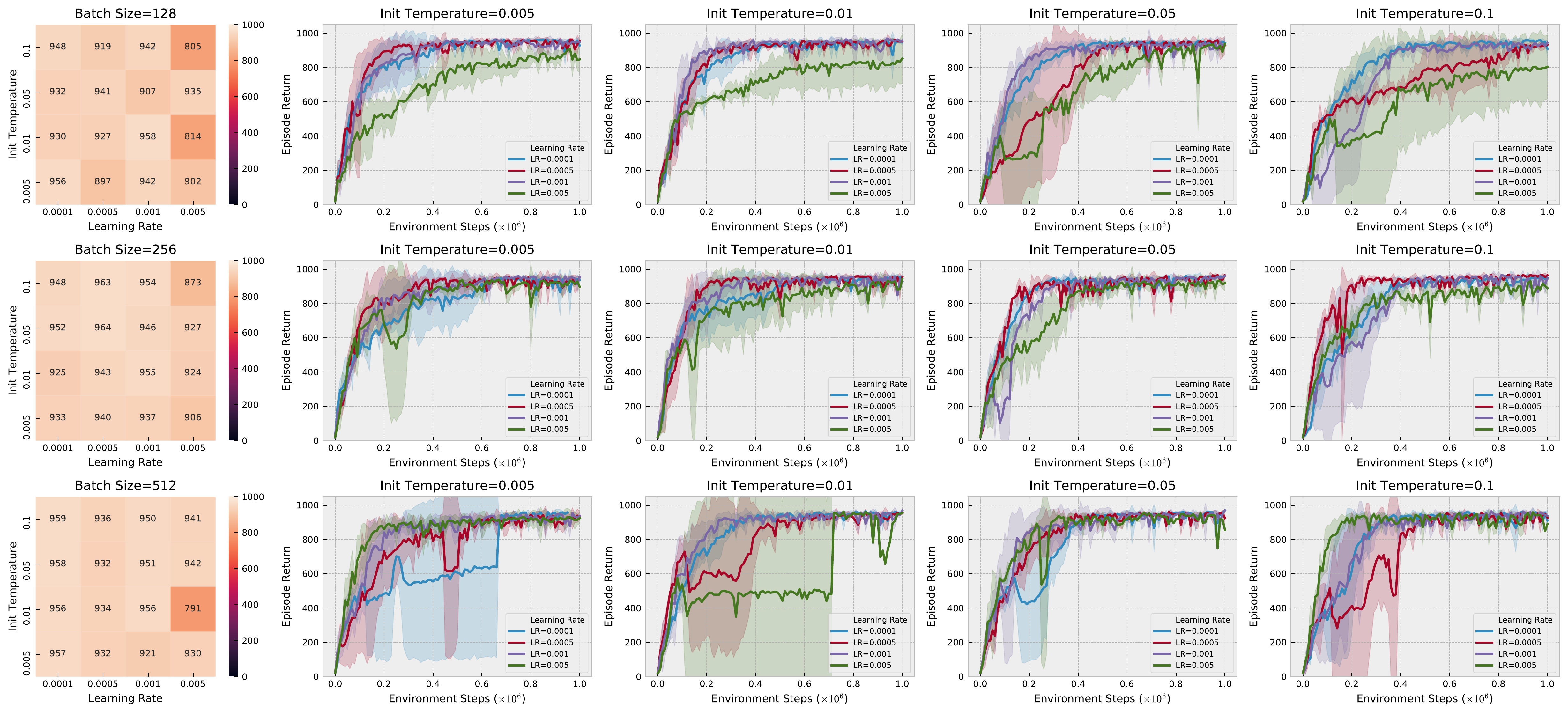}  \label{fig:robust_walker2}}\\
    \subfloat[Cartpole Swingup.]{\includegraphics[width=1.\linewidth]{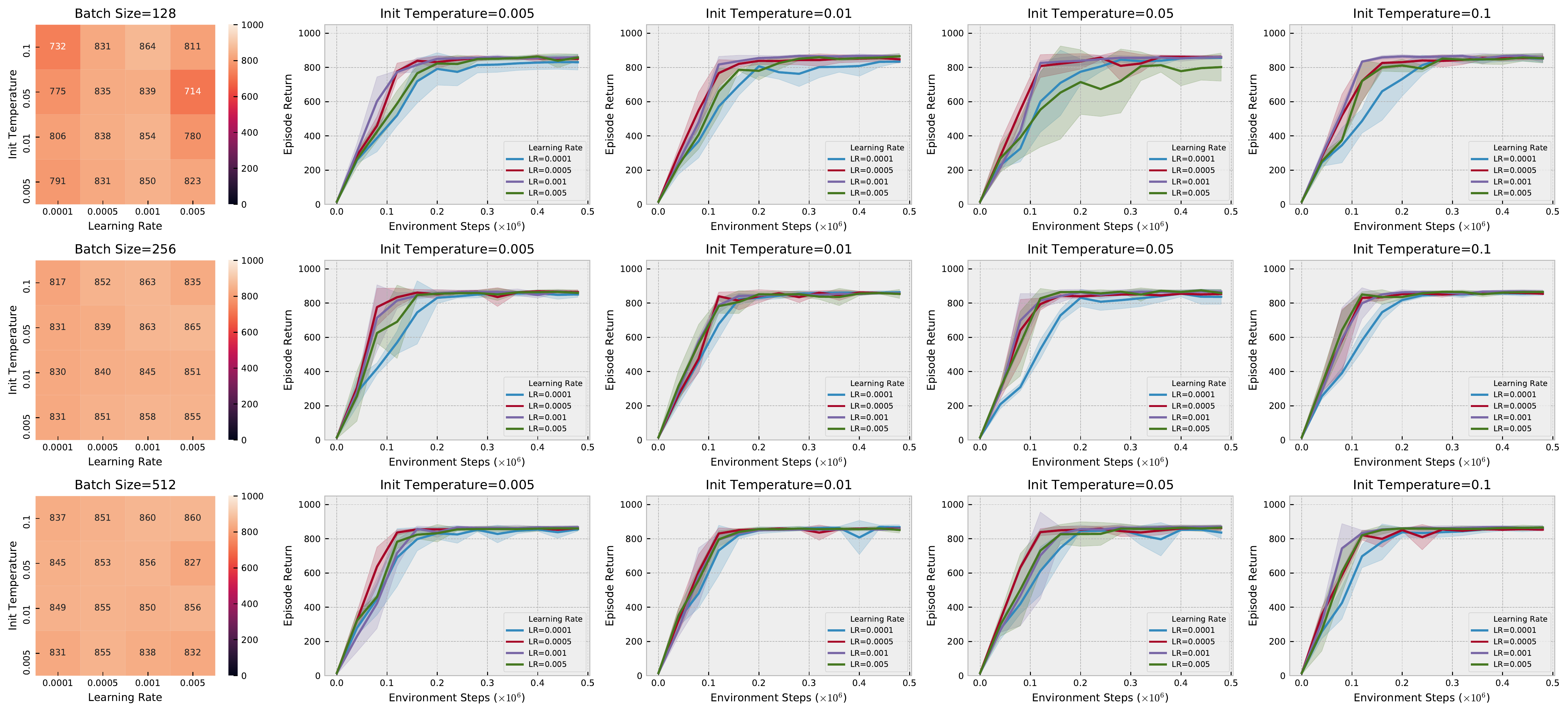}  \label{fig:robust_cartpole2}}\\
    %\vspace{-10pt}
    \subfloat[Finger Spin.]{\includegraphics[width=1.\linewidth]{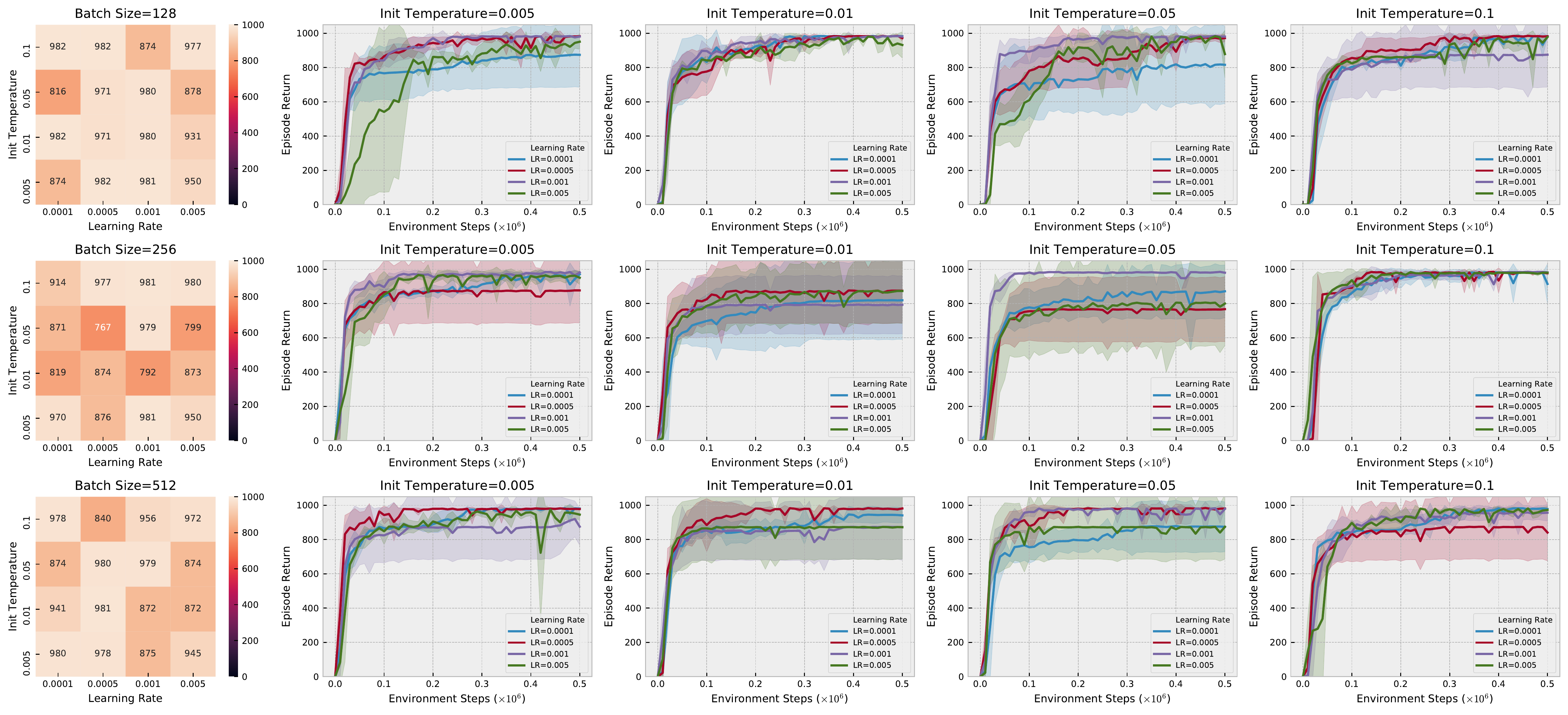}  \label{fig:robust_finger2}}\\
    %\vspace{-10pt}
    \caption{A robustness study of our algorithm (\textbf{DrQ}) to changes in mini-batch size, learning rate, and initial temperature hyper-parameters on three different tasks from \cite{tassa2018dmcontrol}. Each row corresponds to a different mini-batch size. The low variance of the curves and heat-maps shows \drq to be generally robust to exact hyper-parameter settings.} 
    \label{fig:robust_study_full}
\end{figure}

\cref{fig:robust_study_full} shows performance  curves for each configuration as well as a heat map over the mean performance of the final evaluation episodes, similar to~\cite{mnih2016a3c}. Our method demonstrates good stability and is largely invariant to the studied hyper-parameters. We emphasize that for simplicity the experiments in \cref{section:experiments} use the default learning rate of Adam~\cite{kingma2014adam} (0.001), even though it is not always optimal.

\section{Improved Data-Efficient Reinforcement Learning from Pixels}

Our method allows to generate many various transformations from a training observation due to the data augmentation strategy. Thus, we  further investigate whether performing more training updates per an environment step can lead to even better sample-efficiency. Following~\cite{van2019use} we compare a single update with a mini-batch of $512$ transitions with $4$ updates with $4$ different mini-batches of size $128$ samples each. Performing more updates per an environment step leads to even worse over-fitting on some tasks without data augmentation (see~\cref{fig:data_eff_no_aug}), while our method \textbf{DrQ}, that takes advantage of data augmentation, demonstrates improved sample-efficiency (see~\cref{fig:data_eff_aug}).

\begin{figure}[ht!]
    \centering
    \subfloat[Unmodified SAC.]{\includegraphics[width=1.\linewidth]{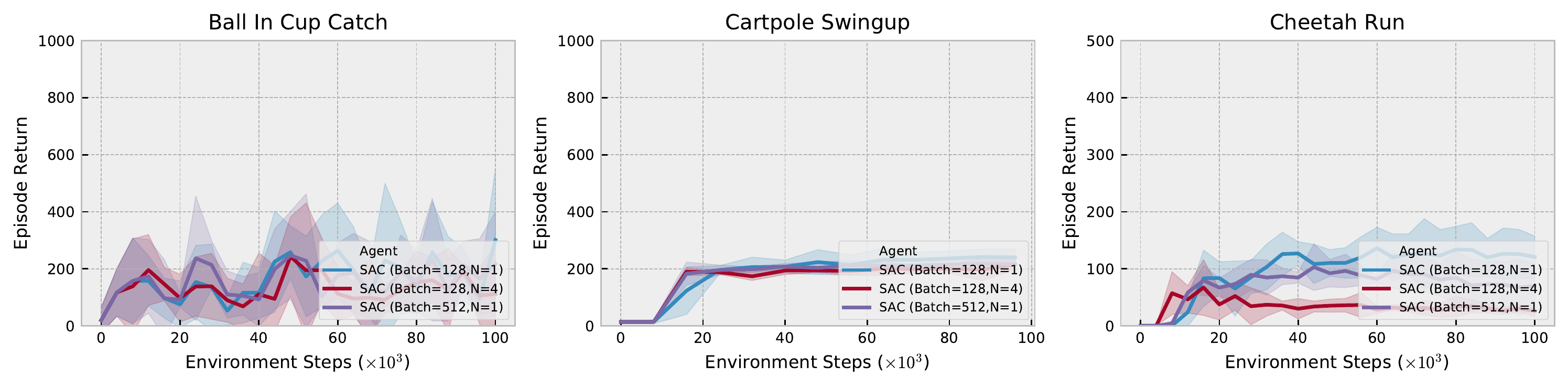}  \label{fig:data_eff_no_aug}}\\
    \subfloat[Our method \textbf{DrQ}.]{\includegraphics[width=1.\linewidth]{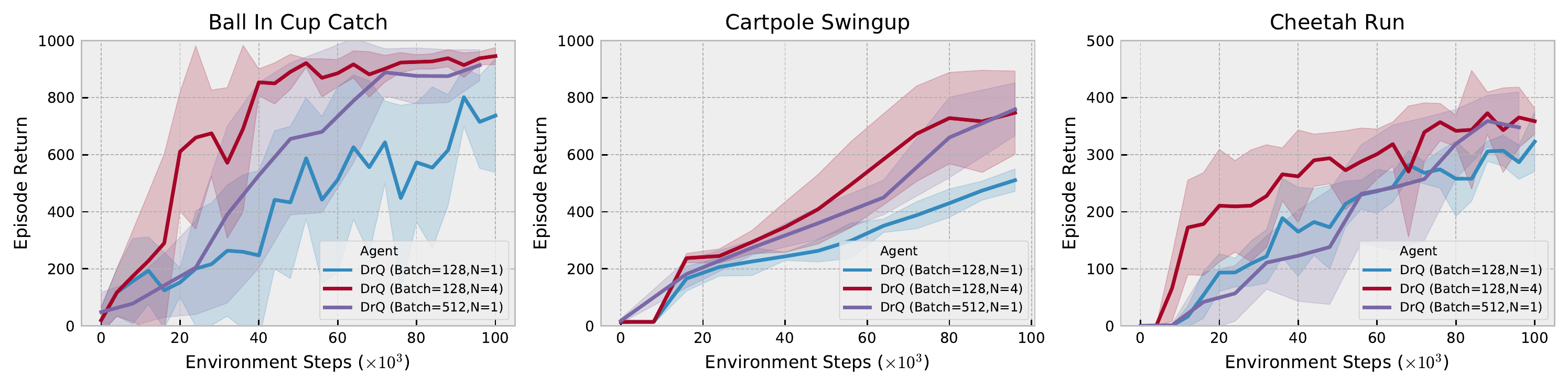} \label{fig:data_eff_aug}}
    \caption{In the data-efficient regime, where we measure performance at 100k environment steps, \textbf{DrQ} is able to enhance its efficiency by performing more training iterations per an environment step. This is because \textbf{DrQ} allows to generate various transformations for a training observation.} 
    \label{fig:data_eff_aug_no_aug}
\end{figure}

\comment{
\section{Experiment with Random Encoder}

We demonstrate that with this data augmentation, even if we freeze the parameters of the convolutional encoder, the method performs better than end-to-end trained encoder without data augmentation. Thus, the regularized by data augmentation the encoder learns features that are crucial for the achieving the state-of-the art performance on the tasks.
}

\end{document}